\documentclass{ps}

\usepackage{amscd,amssymb,graphics}

\usepackage{hyperref}

\hypersetup{
  colorlinks=true,       % false: boxed links; true: colored links
  linkcolor=magenta,          % color of internal links
    filecolor=magenta,      % color of file links
    urlcolor=cyan           % color of external links
}
\newcommand{\R}{{\mathbb R}}
\newcommand{\I}{{\mathbb I}}
\newcommand{\C}{{\mathbb C}}

\newcommand{\ve}{{\varepsilon}}
\def\norm #1{{\left\Vert\,#1\,\right\Vert}}
\newcommand{\abs}[1]{\lvert#1\rvert}

\newcommand{\e}{{\epsilon}}
\newcommand{\N}{{\mathbb N}}

\newcommand{\E}{{\mathbb E}}
\newcommand{\Hg}{{\mathbb H}}

\begin{document}

%%-----------------------------
%%      the top matter
%%-----------------------------
\title{Universal consistency of the $k$-NN rule in metric spaces and Nagata dimension. III}

\runningtitle{$k$-NN rule in metric spaces. III}

\author{Vladimir G. Pestov}\address{Departamento de Matem\'atica, Universidade Federal de Santa Catarina, Trindade, Florian\'opolis, SC, 88.040-900, Brazil (Senior Visiting Professor)}
\secondaddress{Department of Mathematics and Statistics,
       University of Ottawa, 
       Ottawa, ON K1N 6N5, Canada (Emeritus Professor) {\tt vpest283@uOttawa.ca}}
\date{Version as of 16:27 BRT, June 4, 2026}
\begin{abstract}
  We establish the last missing link allowing to describe those complete separable metric spaces $X$ in which the $k$ nearest neighbour classifier is universally consistent, both in combinatorial terms of dimension theory and via a fundamental property of real analysis. 
 The following are equivalent:
  \begin{enumerate}
  \item \label{one}
    The $k$-nearest neighbour classifier is universally consistent in $X$,
  \item \label{two} 
    The strong Lebesgue--Besicovitch differentiation property holds in $X$ for every locally finite\\ Borel measure,
    \item\label{three} $X$ is sigma-finite dimensional in the sense of Jun-Iti Nagata.
  \end{enumerate}
  The equivalence (\ref{two})$\iff$(\ref{three}) was announced by Preiss (1983), though a detailed proof of the implication (\ref{three})$\Rightarrow$(\ref{two}) has only appeared in Assouad and Quentin de Gromard (2006). The implication (\ref{two})$\Rightarrow$(\ref{one}) was established by C\'erou and Guyader (2006). We prove the implication (\ref{one})$\Rightarrow$(\ref{three}).
  We further show that the weak (instead of strong) Lebesgue--Besicovitch property is insufficient for the consistency of the $k$-NN rule, as witnessed, for example, by the Heisenberg group (here we correct a wrong claim made in the previous article (Kumari and Pestov 2024)). A bit counter-intuitively, there is a metric on the real line uniformly equivalent to the usual distance but under which the $k$-NN classifier fails. Finally, another equivalent condition that can be added to the above is the Cover--Hart property:
  \\[.5mm]
\phantom{xxx} 
(4) the error of the $1$-nearest neighbour classifier is asymptotically at most twice as bad as the Bayes error.
\end{abstract}


\maketitle

\section{Introduction}
The $k$ nearest neighbour classification rule is arguably one of the oldest learning algorithms, steeped in real-life practices, and at the same time it is among the learning rules most subjected to rigorous theoretical analysis. It was historically the first learning rule whose universal consistency (in the Euclidean space) was established, in the highly influential 1977 work of Charles J. Stone \cite{stone:77}. (Recall that a learning rule is universally consistent if its expected learning error converges, under any distribution of the labelled data, to the optimal, Bayes, error as the size of the random sample increases.) There is at least one monograph entirely dedicated to the $k$ nearest neighbour method \cite{BD}. At the same time, we are still far from fully understanding the algorithm and there are serious theoretical challenges in the area still to be addressed. The present paper aims to resolve one of the most natural such challenges.

Ample empirical evidence suggests that the efficiency of the $k$-NN algorithm depends on the choice of the domain metric, which may or may not be Euclidean. There is an extensive literature devoted to the choice of a metric based on the data, see the surveys \cite{kulis,BHS}. At the same time, the universal consistency of the variable metric $k$-NN algorithms, such as the Large Margin Nearest Neighbour (LMNN) classifier \cite{WS}, is still an open question.
So one of the most natural problems to begin with would be to determine for which metrics on a given data domain is the $k$-NN classifier universally consistent.

Here a pivotal investigation belongs to C\'erou and Guyader \cite{CG}. First, they have ruled out non-separable metric spaces via a heuristic argument, which was later formalized by Hanneke {\em et al.} \cite{HKSW} to lead to the following alternative: in a non-separable metric space, either the $k$-NN rule is inconsistent (assuming the existence of certain so-called large cardinals), or the $k$-NN consistence completely reduces to the same property in separable subspaces. Thus, only separable metric spaces deserve investigation. And by adding the completeness condition we do not lose in generality (as every metric space can be completed), yet the study becomes technically simpler. Thus, we can and will naturally restrict ourselves to complete separable metric spaces.

The second achievement of the above paper \cite{CG} was to establish the link between real analysis in metric spaces with measure and the consistency of the $k$-NN classifier. Recall that a model of statistical learning in a domain (in our case, metric space) $X$ states that the data is generated in an i.i.d. fashion following a probability measure $\mu$, and the labels of datapoints $x$ are independent Bernoulli variables with variable probability of success $\eta(x)$, where $\eta$ is the regression function on $X$. The main result of \cite{CG} is that if the regression function $\eta$ satisfies the strong Lebesgue--Besicovitch differentiation property with regard to the measure $\mu$, then the $k$-NN classifier is consistent.

In real analysis, the strong Lebesgue--Besicovitch differentiation property (defined a few lines below) is of great importance in various contexts. For the metric spaces, it was thoroughly investigated in particular by David Preiss \cite{preiss83} who in particular has characterized the metric spaces possessing this property under all probability measures in terms of a certain combinatorial parameter of metric geometry, discovered by Nagata \cite{nagata1} in his topological investigations: the metric dimension.

Now we can claim the validity of the following result, where statistical learning, real analysis, and dimension theory all come together.

\begin{thrm}
For a complete separable metric space $\Omega$, the following are equivalent.
\begin{enumerate}
\item The $k$-NN classifier is universally consistent in $\Omega$.
\item For every locally finite Borel measure $\mu$ on $\Omega$, every $L^1(\mu)$-function $f\colon\Omega\to\R$ satisies the strong Lebesgue--Besicovitch differentiation property: for $\mu$-a.e. $x\in\Omega$,
\begin{equation*}
\frac{1}{\mu(B_r(x))}\int_{B_r(x)} f(x^\prime)\,d\mu(x^\prime) \to f(x) ~~\mbox{when $r\to 0$}.
\label{eq:lbs}
\end{equation*}
\item The space $\Omega$ is sigma-finite dimensional in the sense of Nagata.
\end{enumerate}
\label{th:equivalences}
\end{thrm}

Here $B_r(x)$ denotes the open ball of radius $r$ around $x$.
As always, we say that the $k$-NN classifier is universally consistent if for any distribution of labelled data, the misclassification error of the classifier on the i.i.d. labelled data $n$-samples converges to the smallest possible (Bayes) misclassification error when $n,k\to\infty$ and $k/n\to 0$.
Here we assume, like in \cite{CG}, the uniform random distance tie-breaking of the $k$-NN classifier in case of distance ties. 

The last property, that of $\Omega$ being sigma-finite dimensional, is a combinatorial property of metric spaces as defined by Preiss \cite{preiss83} based on the work of Nagata \cite{nagata1,nagata} in topological dimension theory. We devote the entire Section \ref{s:nagata} to treating this property in great detail.

In the above result, only the implication (1)$\Rightarrow$(3) remained unknown, and we establish it in the new result below (Theorem \ref{th:main}). The other implications are known: the equivalence (3) $\iff$ (2) is due to Preiss \cite{preiss83} (the full proof of the implication (3)$\Rightarrow$(2) was worked out in great detail and even for distances more general than metrics by Assouad and Quentin de Gromard \cite{AQ}), while (2)$\Rightarrow$(1) follows from the main result of C\'erou and Guyader \cite{CG} (whose statement was corrected in recent private communication from Fr\'ed\'eric C\'erou \cite{C}). Also, in \cite{CKP} the implication (3)$\Rightarrow$(1) was established directly, in the spirit of the original proof of Stone for the Euclidean case \cite{stone:77}. 

The main new contribution of this paper verifying the implication (1)$\Rightarrow$(3) is the following.

 \begin{thrm}
   Let $\Omega$ be a (complete, separable) metric space that is not sigma-finite dimensional in the sense of Nagata. Then the $k$-NN classifier is not universally consistent in $\Omega$.
   
Moreover, a stronger conclusion holds. Let $k_n\to\infty$ be a sequence of natural numbers with $k_n/n\to 0$. There exist a Borel probability measure $\mu$ on $\Omega$ and a $\mu$-measurable regression function $\eta\colon \Omega\to \{0,1\}$, with $P[\eta(X)=0]=1/2=P[\eta(X)=1]$,
such that the misclassification error of the $k$-NN classifier using $k_{n}$ nearest neighbours in a random i.i.d. $n$-sample following the law $\mu$ with $\eta$ as conditional probability of label $1$ does not converge in probability to zero (which is the Bayes error in this case).
\label{th:main}
 \end{thrm}

Let us stress that the above result we do not impose the usual restriction $k_n\to\infty$. This allows us to obtain a criterion for the $1$-NN classifier later on (Theorem \ref{th:1NN} below).

The result holds under any strategy of distance tie-breaking. The measure $\mu$ built in the proof in essence extends the construction of a measure due to Preiss \cite{preiss79} from the Hilbert space $\ell^2$ to any complete separable metric space that is not sigma-finite dimensional in the sense of Nagata, as defined by Preiss elsewhere \cite{preiss83}. We present the proof in Section \ref{s:measure}.

 In addition to Theorem \ref{th:equivalences}, another consequence of Theorem \ref{th:main} is the observation that the universal consistency of the $k$-NN rule indeed requires the {\em strong}, rather than {\em weak,} Lebesgue--Besicovitch differentiation property. The original theorem of C\'erou and Guyader \cite{CG} affirmed the consistency of the $k$-NN algorithm under the assumption that the regression function satisfies the {\em weak} version of the Lebesgue--Besicovitch property, that is, the convergence in eq. (\ref{eq:lbs}) in measure, rather than almost everywhere. However, it was noted by one of the above authors \cite{C} that the proof implicitly uses the {\em strong} version of the property.
 
 And indeed, now we can indicate examples of complete separable metric spaces and a probability measures on them which satisfy the weak Lebesgue--Besicovitch differentiation property, yet the $k$-NN classifier is not universally consistent. One of them is a metric space constructed by K\"aenm\"aki, Rajala and Suomala \cite{KRS}. Another is the Heisenberg group equipped with the Carnot--Cygan--Koranyi metric, see e.g. \cite{KR,SW,leD} or a relevant discussion and references in \cite{KP}. The Heisenberg group is known to have finite metric dimension in the sense of de Groot \cite{dG}, which, according to Assouad and Quentin de Gromard \cite{AQ}, implies it has the weak Lebesgue--Besicovitch differentiation property for every measure.

 In \cite{KP}, we assumed the validity of the result from \cite{CG} that the weak Lebesgue--Besicovitch differentiation property suffices for the weak consistency of the $k$-NN classifier, and erroneously claimed that the $k$-NN classifier is universally consistent in the Heisenberg group.  Our Theorem \ref{th:main} implies this is not true, as the Heisenberg group is not sigma-finite dimensional in the sense of Nagata \cite{KR,SW}. This is the content of Section \ref{s:weak}.

 Another interesting consequence of the main result, given in Section \ref{s:R}, is that every complete separable metric space of uncountable cardinality admits a uniformly equivalent metric under which the $k$-NN classifier is not universally consistent -- which looks rather counter-intuitive in the particular case of the real line.

 The equivalent conditions in Theorem \ref{th:equivalences} can be also expressed in terms of the $1$-nearest neighbour classifier. We are stating the result separately so as not to overload the original theorem. In the classical theory of nearest neighbour classifier in the Euclidean space, it is known that the error of the $1$-NN classifier, asymptotically, is at most twice as bad as the Bayes error \cite{CH,stone:77}. This property is characteristic of the sigma-finite dimensional metric spaces as well.

 \begin{thrm}
 Let $X$ be a complete separable metric space.  The following two conditions are equivalent between themselves and to all the conditions in Theorem \ref{th:equivalences}:
   \begin{enumerate}
   \item[(4)] Under every distribution of the labelled data on $X\times\{0,1\}$, the misclassification error of the $1$-nearest neighbour rule, asymptotically as the size of an i.i.d. labelled sample goes to infinity, does not exceed twice the Bayes error.
     \item[(4$^\prime$)] Under every distribution of the labelled data on $X\times\{0,1\}$ whose Bayes error is zero (that is, the regression function is deterministic), the $1$-nearest neighbour rule is consistent when the size of an i.i.d. labelled sample goes to infinity.
   \end{enumerate}
   \label{th:1NN}
 \end{thrm}

 This result is established in the concluding Section \ref{s:1NN}.

This is the third and last article in a series \cite{CKP,KP}, but it is largely self-contained.
We set things up in Section \ref{s:prelim}. In Section \ref{s:nagata} we discuss the Nagata dimension, and the remaining Sections contain proofs.

This note was prompted by a private message \cite{C} from the first author of the paper \cite{CG} correcting the statement of their main theorem. For this, I am most grateful to Fr\'ed\'eric C\'erou, as well as to L\'aszl\'o Gy\"orfi, Aryeh Kontorovich, Harro Walk, and Roi Weiss for stimulating discussions, without which this paper would not have appeared. The author is equally grateful to the anonymous journal referee for two very substantial consecutive reports that have led to the first two versions of the article being considerably reworked.

\section{Preliminaries\label{s:prelim}}
\subsection{General setting}
The standard model for supervised binary classification in  statistical learning theory starts with the domain, which is a measurable space, that is, a set, $\Omega$, equipped with a sigma-algebra of subsets $\mathcal A$. (We use throughtout the rest of the paper the letter $\Omega$, reserving the letter $X$ for a random element.) The theory is only developed to any extent in the particular case where $\Omega$ is a standard Borel space. In other words, there is a metric $\rho$ on $\Omega$ making it a complete separable metric space, and such that
the sigma-algebra $\mathcal A$ consists of the Borel subsets, that is, $\mathcal A$ is the smallest sigma-algebra containing all the open sets. It is a fundamental result of descriptive set theory that  any two standard Borel spaces of the same cardinality are isomorphic between themselves, and any such space of uncountable cardinality is isomorphic, as a measurable space, to the unit interval (Th. 15.6 in \cite{kechris}). In our case, $\Omega$ will be a complete separable metric space, thus, in particular, a standard Borel space.

A classifier in a domain $\Omega$ is a measurable mapping $T\colon\Omega\to\{0,1\}$.
A finite sample in $\Omega$ is a sequence $(x_1,x_2,\ldots,x_n)$, $x_i\in\Omega$, $n\in\N_+$. We will consider the case of binary labels taking values $0$ or $1$. A labelled sample is a finite sequence of elements of the labelled domain, $\Omega\times\{0,1\}$, written $\sigma=(x_1,y_1,\ldots,x_n,y_n)$, $x_i\in\Omega$, $y_i\in\{0,1\}$.
A learning rule for $\Omega$ is a countable sequence $g=(g_n)_{n=1}^{\infty}$ of mappings, associating to each labelled sample a classifier, $g_n(\sigma)\colon\Omega\to\{0,1\}$. Each $g_n$ can be viewed as a mapping
\[g_n\colon \Omega^n\times\{0,1\}^n\times\Omega\to \{0,1\},\]
and we request those mappings to be measurable (with regard to the product sigma-algebra).

It is further assumed that the labelled pairs $(X,Y)$ are random elements of $\Omega\times\{0,1\}$ following a certain (unknown) probability measure $\tilde\mu$. The samples are i.i.d. sequences of such random elements following the same law.

It is usually convenient to reinterpret $\tilde\mu$ in a different way. Denote $\pi\colon\Omega\times\{0,1\}\to\Omega$ the first coordinate projection, $\pi(x,y)=x$. The direct image of $\tilde\mu$ under this projection is a probability measure on $\Omega$, $\mu=\tilde\mu\circ\pi^{-1}$. This $\mu$ is the law of unlabelled datapoints. Now restrict $\tilde\mu$ to the subspace $\Omega\times\{1\}$ and project this restriction under $\pi$ onto $\Omega$. Denote this measure $\mu_1$: $\mu_1(A) = \tilde\mu(A\times\{1\})$. Since $\mu_1$ is clearly absolutely continuous with regard to $\mu$, there exists for $\mu$-a.e. $x$ the Radon-Nykodym derivative:
\begin{align*}
  \eta(x)&=\frac{d\mu_1}{d\mu}(x)\in [0,1] \\
  &=P[Y=1\mid X=x].
  \end{align*}
This is the conditional probability for a point $x$ to be labelled $1$. In this context, $\eta$ is referred to as the regression function. Since the measure $\tilde\mu$ is in its turn completely recoverable from the pair $(\mu,\eta)$, it is an often technically preferable way to encode the learning problem.

The learning (or generalization) error of a classifier, $T\colon\Omega\to\{0,1\}$ is the quantity
\begin{align*}
  \ell_{\tilde\mu}(T)&=\tilde\mu\{(x,y)\in\Omega\times\{0,1\}\colon T(x)\neq y\} \\
  &= P[T(X)\neq Y].
\end{align*}
The infimum of errors of all classifiers (which is in fact the minimum) is called the Bayes error:
\begin{align*}
  \ell^{\ast}_{\tilde\mu} &= \inf_{T\colon\Omega\to\{0,1\}}\ell_{\tilde\mu}(T)\\
  &= \int_{\Omega}\min\{\eta(x),1-\eta(x)\}\,d\mu(x).
\end{align*}
A classifier is called a Bayes classifier if its error equals the Bayes error. Those are exactly all classifiers satisfying $\mu$-a.e. the property
\[T_{bayes}(x) = \begin{cases} 1,&\mbox{ if }\eta(x)>\frac 12, \\
  0,&\mbox{ if }\eta(x)<\frac 12.
\end{cases}
\]
The Bayes error is zero if and only if the regression function is deterministic, that is, for $\mu$-a.e. $x$, $\eta(x)\in\{0,1\}$. 

Given a learning rule $g$, in the presence of a probability measure $\tilde\mu$ the error of $g_n$ is a random variable:
%%expected error of classifiers of the form $g_n(\sigma)$:
\[\ell_{\mu}(g_n)=\ell_{\tilde\mu}(g_n(\sigma)),~~\sigma\sim\tilde\mu^n.\]
The learning rule $g$ is called consistent under $\tilde\mu$ if the error of $g_n$ converges to the Bayes error in probability (equivalently, in expectation) as $n\to\infty$:
\[\ell_{\tilde\mu}(g_n)\overset p\to\ell^{\ast}_{\tilde\mu}.\]
The rule $g$ is universally consistent on the measurable space $\Omega$ if it is consistent under every probability measure on the measurable space $\Omega\times\{0,1\}$.

Sometimes the consistent rules are called weakly consistent, to distinguish them from strongly consistent rules, where the convergence to the Bayes error is along a.e. every sample path. However, the strong consistency of learning rules appears to be a rare phenomenon (see \cite{KP}, Subsection 2.3).

We are studying the $k$ nearest neighbour rule for supervised binary classification.
The $k$-NN classifier assigns the label $0$ or $1$ to every point of $\Omega$ according to the majority vote among the labels of the $k$ nearest neighbours in a random labelled $n$-sample. The nearest neighbours are chosen with regard to the metric $\rho$ on $\Omega$. There are two tie problems to resolve.

The first is the distance tie. Denote $r_k(x)=r_k(\sigma,x)$ the smallest radius of a closed ball around $x$ containing $k$ elements of the random sample $\sigma$:
\[r_k(x)=\min\left\{r\geq 0\colon \sharp\{i\colon \rho(x,x_i)\leq r_k \}\geq k\right\}.\]
It may happen that there are strictly more than $k$ points at a distance $\leq r_k(x)$ from $x$. In this case, one needs an agreed algorithmic procedure to select the $k$ nearest neighbours of $x$: one selects all elements of the open $r_k$-sphere plus the necessary number of those sitting on the boundary. The usual choice is a random uniform tie-breaking. One chooses an auxiliary i.i.d. random sequence $\xi_i$ following some non-atomic probability distribution on the real line, independent of the data, and when having to choose between a few nearest neighbours at the same distance $r_k(x)$ from $x$, one gives preference to the indices $i$ for which $\xi_i$ is smaller.

The algorithm chosen for distance tie-breaking is unimportant for the main result of this article (Theorem \ref{th:main}).
However, for the implication (2)$\Rightarrow$(1) established in \cite{CG}, it is assumed to be a random uniform tie-breaking.

The second kind of tie is the voting tie when $k$ is even and exactly half among the $k$ nearest neighbours of $x$ are labelled $0$ and other half, $1$. Here the tie-breaking can be any, as it can be proved that it has no influence on the consistency of the $k$-NN classifier. Most often, one breaks the voting tie by choosing the label $1$.

The $k$-NN learning rule is a family of classifiers of the above form, under a choice of a sequence of values $k=k_n$ satisfying $k\to\infty$ and $k/n\to 0$ as $n\to\infty$.

Detailed presentations of the $k$-NN classifier in the metric space setting can be found in numerous sources, including \cite{CG,CKP,KP}.

Now we will survey the previously known implications in Theorem  \ref{th:equivalences}.

\subsection{Equivalence (3)$\iff$(2)\label{subs:pag}}
The equivalence (3)$\iff$(2) of Theorem \ref{th:equivalences} is due to Preiss who has sketched the proofs in a short note \cite{preiss83} in a rather terse way, mostly paying attention to the implication (2)$\Rightarrow$(3). The proof of the implication (3)$\Rightarrow$(2) was only worked out in great detail (and in a more general setting) by Assouad and Quentin de Gromard (see \cite{AQ}, Thm. 4.4.(n6)), and this is the only implication we will rely on to conclude the equivalence of all three conditions.

\subsection{Implication (2) $\Rightarrow$ (1)\label{ss:21}} This implication follows from the result of C\'erou and Guyader \cite{CG}.

\begin{dfntn}
  A complete separable metric space $\Omega$ equipped with a locally finite Borel measure
  %% (equivalently: just a Borel probability measure)
  $\mu$ satisfies the {\em strong} {\em Lebesgue--Besicovitch differentiation property} if for every $L^1(\mu)$-function $f\colon\Omega\to\R$ and for $\mu$-a.e. $x$,
 \begin{equation}
\lim_{r\downarrow 0}\frac{1}{\mu(B_r(x))}\int_{B_r(x)} f(x^\prime)\,d\mu(x^\prime) = f(x).
\label{eq:slb}
 \end{equation}
 The metric space $\Omega$ with measure as above satisfies the {\em weak Lebesgue--Besicovitch differentiation property} if for every function $f$ as above,
  \begin{equation}
\frac{1}{\mu(B_r(x))}\int_{B_r(x)} f(x^\prime)\,d\mu(x^\prime) \to f(x)\mbox{ as }r\downarrow 0,
\label{eq:wlb}
\end{equation}
where the convergence is in measure, that is, for each $\e>0$,
\[\mu\left\{x\in\Omega\colon 
\left\vert\frac{1}{\mu(B_r(x))}\int_{B_r(x)} f(x^\prime)\,d\mu(x^\prime) - f(x)
\right\vert >\e\right\}\to 0\mbox{ when } r\downarrow 0.\]
\end{dfntn}

The original theorem in \cite{CG} stated that for every complete separable metric space equipped with a probability measure $\mu$ and a regression function $\eta$ satisfying the {\em weak} Lebesgue--Besicovitch differentiation property the $k$-NN classifier is consistent.

However, one of the authors (Fr\'ed\'eric C\'erou) has kindly pointed out to us \cite{C} that the very last assertion in the proof of the main result of \cite{CG}, at the bottom of p. 345, requires the {\em strong} Lebesgue--Besicovitch property. We reproduce the relevant part of the proof.

\begin{quote}
  ``Fix $\varepsilon>0$, then for every $\delta_0>0$
  \begin{align*}
    P\left(\mu\left(B_{X,d_{k+1}}\right)^{-1}\int_{B_{X,d_{k+1}}}\left\vert\eta-\eta(X) \right\vert\,d\mu >\varepsilon \right) &\leq P\left(d_{(k+1)(X)}\geq\delta_0 \right) +\\ & \sup_{0\leq\delta\leq\delta_0}P\left(\left(\mu(B_{X,\delta})) \right)^{-1}\int_{B_{X,\delta}}\left\vert\eta-\eta(X) \right\vert\,d\mu>\varepsilon \right).
  \end{align*}
  Now, the first term goes to zero thanks to Cover and Hart's result and the second one also thanks to the Besicovitch condition.''
\end{quote}

\vskip .2cm
Here $d_{k+1}$ is the radius of the smallest ball containing $k+1$ nearest neighbours of a random element $X$, in our notation it would be $r_{k+1}$.
To better see the problem, let us rewrite the inequality measure-theoretically, denoting $d_{k+1}$ by $r(x)$:
\begin{align*}
  \mu\left\{x\colon  \left\vert \frac{1}{\mu(B_{r(x)}(x))}\int_{B_{r(x)}(x)} f(x^\prime)\,d\mu(x^\prime) - f(x)\right\vert>\varepsilon \right\} &\leq
\mu\left\{x\colon r\geq \delta_0 \right\} + \\ &
\sup_{0\leq\delta\leq\delta_0}\mu\left\{x\colon \left\vert \frac{1}{\mu(B_{\delta}(x))}\int_{B_\delta(x)} f(x^\prime)\,d\mu(x^\prime) - f(x)\right\vert>\varepsilon \right\}.
\end{align*}
If $r(x)=d_{k+1}$ were a constant, it would be perfectly correct. But the radius depends on $x$, and one can imagine a function $f$ for which at every $x$ the absolute value between the braces spikes up over $\e$ for some suitable $r=r(x)<\delta_0$, different for each $x$, and yet it happens on an ever smaller set of $x$ as $r\downarrow 0$, so that the convergence in probability in Eq. (\ref{eq:wlb}) still takes place. To guarantee the inequality, in the last term we must swap the supremum and the measure sign:
\begin{align*}
  \mu\left\{x\colon  \left\vert \frac{1}{\mu(B_{r(x)}(x))}\int_{B_{r(x)}(x)} f(x^\prime)\,d\mu(x^\prime) - f(x)\right\vert>\varepsilon \right\} &\leq
\mu\left\{x\colon r\geq \delta_0 \right\} + \\ &
\mu\left\{x\colon \sup_{0\leq\delta\leq\delta_0}\left\vert \frac{1}{\mu(B_{\delta}(x))}\int_{B_\delta(x)} f(x^\prime)\,d\mu(x^\prime) - f(x)\right\vert>\varepsilon \right\}.
\end{align*}

In this form, the last term converges to zero when $\delta_0\to 0$ if (and only if) the function $\eta$ has the {\em strong} Lebesgue--Besicovitch property.

(This author must confess that this subtlety had escaped his attention during what he thought was a careful reading of the paper \cite{CG}.)

Thus, the corrected form of the result is this.

\begin{thrm}[C\'erou and Guyader \cite{CG}]
  Let $\Omega$ be a complete separable metric space equipped with a probability measure, $\mu$, and a $\mu$-measurable regression function $\eta\colon\Omega\to [0,1]$. If $\eta$ satisfies the strong Lebesgue--Besicovitch differentiation property with regard to $\mu$, then the $k$-NN classifier is consistent with the learning problem given by $(\mu,\eta)$ whenever $n,k\to\infty$ and $k/n\to 0$.
\end{thrm}

In particular, this validates the implication (2)$\Rightarrow$(1) in Thm. \ref{th:main}. In Section \ref{s:weak} we will give examples showing that indeed the original statement of the theorem, and not just the proof, was not right and the weak Lebesgue--Besicovitch differentiation property does not in general imply the $k$-NN rule consistency.

\subsection{Implication (3) $\Rightarrow$ (1)} We also note that this implication was established directly in our previous work \cite{CKP} in the same spirit as the original result of Charles J. Stone for the $k$-NN classifier in the Euclidean space \cite{stone:77}, though with some interesting geometrical complications.

\subsection{Implication (1) $\Rightarrow$ (3)} This implication in Theorem \ref{th:equivalences} follows from our Theorem \ref{th:main}.

\section{Nagata dimension\label{s:nagata}}

This Section is largely (though not entirely) a compilation of known results. Since they are scattered in the literature \cite{AQ,CKP,KP,nagata1,nagata,nagata_open,ostrand,preiss83} and for some of them the proofs are missing, we are giving a self-contained presentation.

Recall that a family $\gamma$ of subsets of a set $\Omega$ has multiplicity $\leq\delta$ if every intersection of more than $\delta$ different elements of $\gamma$ is empty. In other words,
\[\forall x\in\Omega,~~\sum_{V\in\gamma}\chi_V(x)\leq\delta,\]
where $\chi_V$ is the indicator function of $V$.

\begin{dfntn} A metric subspace $X$ of a metric space $\Omega$ has {\em Nagata dimension $\leq\delta\in\N$ on the scale $s\in (0,+\infty]$ inside of} $\Omega$ if every finite family of closed balls in $\Omega$ with centres in $X$ and radii $<s$ admits a subfamily of multiplicity $\leq\delta+1$ in $\Omega$ which covers all the centres of the original balls. The subspace $X$ has a finite Nagata dimension in $\Omega$ if $X$ has finite dimension in $\Omega$ on some scale $s>0$. Notation for the smallest $\delta$ with the above property: $\dim^s_{Nag}(X,\Omega)$ or sometimes simply $\dim_{Nag}(X,\Omega)$.
\label{d:dimnagata}
\end{dfntn}

Following Preiss \cite{preiss83}, let us call a family of balls {\em disconnected} if the centre of each ball does not belong to any other ball from the family. Here is a reformulation of the above definition.

\begin{prpstn}  
For a subspace $X$ of a metric space $\Omega$, one has
\[\dim^s_{Nag}(X,\Omega) \leq \delta\]
if and only if every disconnected family of closed balls in $\Omega$ of radii $<s$ with centres in $X$ has multiplicity $\leq \delta+1$.
\label{ex:famdesconexa}
\end{prpstn}

\begin{proof}
{\em Necessity.} Let $\gamma$ be a disconnected finite family of closed balls in $\Omega$ of radii $<s$ with centres in $X$. Since by assumption $\dim^s_{Nag}(X,\Omega) \leq \delta$, the family $\gamma$ admits a subfamily of multiplicity $\leq\delta+1$ covering all the original centres. But the only subfamily that contains all the centres is $\gamma$ itself.

{\em Sufficiency.} 
Let $\gamma$ be a finite family of closed balls in $\Omega$ having radii $<s$ and centres in $X$.
Denote $C$ the set of centres of those balls. 
Among all the disconnected subfamilies of $\gamma$ (which exist, e.g., each family containing just one ball is such) choose one, $\gamma^\prime$, having the maximal cardinality of the set $C\cap \cup\gamma^\prime$. We claim that $C\subseteq\cup\gamma^\prime$ (which will finish the argument because the multiplicity of $\gamma^{\prime}$ is bounded by $\delta+1$). Indeed, assume there is a ball, $B\in\gamma$, whose centre, $c\in C$, does not belong to $\cup \gamma^\prime$. Remove from $\gamma^\prime$ all the balls with centres in $B\cap C$ and add $B$ instead. The new family, $\gamma^{\prime\prime}$, is still disconnected and covers $\left(C\cap\cup\gamma^\prime\right)\cup\{c\}$, which contradicts the assumed maximality property of $\gamma^\prime$. 
\end{proof}

In the definition \ref{d:dimnagata}, as well as in the proposition \ref{ex:famdesconexa}, it is convenient to replace closed balls with open ones.

\begin{prpstn} 
For a subspace $X$ of a metric space $\Omega$, the following are equivalent.
\begin{enumerate}
\item\label{equiv:1} $\dim^s_{Nag}(X,\Omega) \leq\delta$,
\item\label{equiv:2} every finite family of open balls in $\Omega$ having radii $<s$ with centres in $X$ admits a subfamily of multiplicity $\leq\delta+1$ in $\Omega$ which covers all the centres of the original balls,
\item\label{equiv:4} every disconnected family of open balls in $\Omega$ of radii $<s$ with centres in $X$ has multiplicity $\leq\delta+1$.
\end{enumerate}
\label{p:clopen}
\end{prpstn}

\begin{proof}
  $(\ref{equiv:1})\Rightarrow(\ref{equiv:2})$: Let $\gamma$ be a finite family of open balls in $\Omega$ with centres in $X$ and of radii $<s$.
  For every element $B\in\gamma$ and each $k\geq 2$, form a closed ball $B_k$
  as having the same centre and radius $r(1-1/m)$, where $r$ is the radius of $B$. Thus, $B=\cup_{k=2}^{\infty}B_k$. Select recursively a chain of subfamilies 
\[\gamma\supseteq \gamma_1\supseteq\gamma_2\supseteq \ldots \supseteq \gamma_k\supseteq \ldots\]
with the properties that for each $k$, the family of closed balls $B_k$, $B\in\gamma_k$ has multiplicity $\leq\delta+1$ in $\Omega$ and covers all the centres of the balls in $\gamma$. Since $\gamma$ is finite, starting with some $k$, the subfamily $\gamma_k$ stabilizes, and now it is easy to see that the subfamily $\gamma_k$ itself has the multiplicity $\leq\delta+1$, and still covers all the original centres.

$(\ref{equiv:2})\Rightarrow(\ref{equiv:4})$:
Same argument word for word as in the proof of necessity in proposition \ref{ex:famdesconexa}.

$(\ref{equiv:4})\Rightarrow(\ref{equiv:1})$: Let $\gamma$ be a disconnected family of
closed balls in $\Omega$,
having radii $<s$ and centres in $X$. For each $B\in\gamma$ and $\ve>0$, denote $B_{\ve}$ an open ball
concentric with $B$ and of the radius $r+\ve$, where $r$ is the radius of $B$.
For a sufficiently small $\ve>0$, the family $\{B_\ve\colon B\in\gamma\}$ is disconnected, and its radii are all strictly less than $s$, therefore this family has multiplicity $\leq\delta+1$ by assumption. The same follows for $\gamma$. Now use proposition \ref{ex:famdesconexa} to conclude.
\end{proof}

\begin{prpstn}
Let $X$ be a subspace of a metric space $\Omega$, satisfying $\dim^s_{Nag}(X,\Omega)\leq \delta$. Then $\dim^s_{Nag}(\bar X,\Omega)\leq \delta$, where $\bar X$ is the closure of $X$ in $\Omega$.
\label{p:closurenagata}
\end{prpstn}

\begin{proof}
  Let $\gamma$ be a finite disconnected family of open balls in $\Omega$ of radii $<s$, centred in $\bar X$. Let $y\in\Omega$, and let $\gamma^\prime$ consist of all balls in $\gamma$ containing $y$. Choose $\ve>0$ so small that the open $\ve$-ball around $y$ is contained, as a subset, in every element of $\gamma^\prime$. For every open ball $B\in\gamma^\prime$, denote $y_B$ the centre and $r_B$ the radius. We can additionally assume that $\ve<r_B$.
  Denote $B^\prime$ an open ball of radius $r_B-\ve>0$, centred at a point $x_B\in X$ satisfying $d(x_B,y_B)<\ve/2$. Note that $y\in B^\prime$ by the triangle inequality. Again by the triangle inequality, the family $\{B^\prime\colon B\in\gamma^\prime\}$ is disconnected, and it has radii $<s$. Therefore, $y$ only belongs to $\leq\delta+1$ balls $B^\prime$, $B\in\gamma^\prime$. Consequently, the cardinality of $\gamma^\prime$ is bounded by $\delta+1$.
\end{proof}

\begin{prpstn}
If $X$ and $Y$ are two subspaces of a metric space $\Omega$, having finite Nagata dimension in $\Omega$ on the scales $s_1$ and $s_2$ respectively, then $X\cup Y$ has a finite Nagata dimension in $\Omega$, with $\dim_{Nag}(X\cup Y,\Omega)\leq \dim_{Nag}(X,\Omega)+\dim_{Nag}(Y,\Omega)+1$, on the scale $\min\{s_1,s_2\}$.
\label{p:uniaonagata}
\end{prpstn}

\begin{proof}
  Given a finite family of balls $\gamma$ in $\Omega$ of radii $<\min\{s_1,s_2\}$ centred in elements of $X\cup Y$, represent it as $\gamma=\gamma_X\cup\gamma_Y$, where the balls in $\gamma_X$ are centered in $X$, and the balls in $\gamma_Y$ are centred in $Y$. Select a subfamily $\gamma_X^{\prime}$ of multiplicity $\dim_{Nag}(X,\Omega)+1$ containing all centres of balls in $\gamma_X$, and a subfamily $\gamma_Y^{\prime}$ of multiplicity $\dim_{Nag}(Y,\Omega)+1$ containing all centres of balls in $\gamma_Y$, and form their union. It has multiplicity $\leq \dim_{Nag}(X,\Omega)+\dim_{Nag}(Y,\Omega)+2$ and covers all the original centres, whence the result follows.
\end{proof}

\begin{rmrk}
 Adding $1$ to the sum of dimensions is unavoidable, as seen from the example $\Omega_1=\{0,1\}$, $\Omega_2=\{1,2\}$ with the usual distance on the real line.
  \end{rmrk}

\begin{dfntn}[Preiss \cite{preiss83}]
  A metric space $\Omega$ is {\em sigma-finite dimensional in the sense of Nagata} if $\Omega=\cup_{i=1}^{\infty}X_n$, where every subspace $X_n$ has finite Nagata dimension in $\Omega$ on some scale $s_n>0$ (where the scales $s_n$ are possibly all different).
  \label{d:sfd}
\end{dfntn}

\begin{rmrk}
  Due to Proposition \ref{p:closurenagata}, in the above definition we can assume the subspaces $X_n$ to be closed in $\Omega$. By Proposition \ref{p:uniaonagata}, we can assume $X_n$ to form an increasing chain. 
\end{rmrk}

\begin{rmrk}
  The Baire category theorem implies that every sigma-finite dimensional complete metric space $\Omega$ contains a non-empty open subset $X$ having a finite metric dimension in $\Omega$.
  \label{r:baire}
  \end{rmrk}

\begin{rmrk}
  Note that the notion of a subset $X$ of a metric space $\Omega$ being finite-dimensional as studied above is relative, that is, it is a property of the pair $(X,\Omega)$ rather than of $X$ itself.
\end{rmrk}

\begin{rmrk}
  A metric space $\Omega$ is said to have finite Nagata dimension if it has finite Nagata dimension in itself, that is, as a pair $(\Omega,\Omega)$. In this case we write $\dim_{Nag}(\Omega)$.
  The Nagata--Ostrand theorem \cite{nagata1,nagata,ostrand} states that the Lebesgue covering dimension of a metrizable topological space is the smallest Nagata dimension of a compatible metric on the space (and in fact this is true on every scale $s>0$). This is the historical origin of the concept of the Nagata metric dimension.
\end{rmrk}

\begin{prpstn}
  Every separable metric space $\Omega$ that is not sigma-finite dimensional (in itself) in the sense of Nagata contains a non-empty closed subspace $\Upsilon$ none of whose non-empty relatively open subsets has finite Nagata dimension in $\Omega$.
  \label{p:yx}
\end{prpstn}

\begin{proof}
  Let us say, slightly extending the definition \ref{d:sfd} of Preiss, that a subset $A$ of $\Omega$ is sigma-finite dimensional in $\Omega$ if one can represent $A$ as the union of a countable family of subsets $A_n$ each of which is finite-dimensional in $\Omega$ on some strictly positive scale.
  
Denote $\gamma$ the family of all non-empty open subsets of $\Omega$ that are sigma-finite dimensional in $\Omega$. Let $O$ be the union of $\gamma$. If it happens that $O=\emptyset$, there is nothing left to prove, so we may as well assume $O\neq\emptyset$. Since $O$ is a separable metric space, it is Lindel\"of, and one can select a countable subfamily of $\gamma$ whose union is $O$ (\cite{engelking}, Thm. 3.8.1). It follows that $O$ itself is sigma-finite dimensional in $\Omega$, so it is the largest open subset that is sigma-finite dimensional in $\Omega$.

Denote $\Upsilon=\Omega\setminus O$. Since $\Omega$ is not sigma-finite dimensional in itself, the subspace $\Upsilon$ is non-empty, and it is also closed. Let now $V$ be a non-empty relatively open subset of $\Upsilon$. Thus, there exists an open $U\subseteq \Omega$ with $U\cap \Upsilon=V$. Suppose that $V$ is sigma-finite dimensional in $X$.
Since $Y$ is closed, $U\setminus V= U\setminus Y$ is an open subset of $\Omega$, in fact of $O$, hence $U\setminus V$ is sigma-finite dimensional in $\Omega$. It follows that $U=V\cup (U\setminus V)$ is an open sigma-finite dimensional subset of $\Omega$ that is not contained in $O$, in contradiction to the maximality of $O$ with these properties. We conclude: every non-empty relatively open subset of $\Upsilon$ is not sigma-finite dimensional in $\Omega$, in particular, has infinite Nagata dimension in $\Omega$.
\end{proof}

\begin{rmrk}
  In \cite{preiss83}, Preiss seems to imply (p. 567, lines 3-4) that every {\em complete} separable metric space $\Omega$ that is not sigma-finite dimensional (in itself) in the sense of Nagata contains a non-empty closed subspace $\Upsilon$ containing no non-empty (relatively) open subsets of finite Nagata dimension in $\Upsilon$ proper, but I am unable to verify this claim. The purported proof of it that my student Sushma Kumari and I had come up with and she included in her Ph.D. thesis (Lemma 1.4.2, p. 35 in \cite{kumari}) appears deficient.
  \end{rmrk}

\begin{xmpl} 
A metric space has Nagata dimension zero on the scale $+\infty$ if and only if it is non-archimedian, that is, satisfies the strong triangle inequality, $d(x,z)\leq \max\{d(x,y),d(y,z)\}$.
\label{ex:nonarchimedian}
\end{xmpl}

\begin{xmpl}
  It is easy to see, especially using Prop. \ref{ex:famdesconexa}, that $\dim_{Nag}(\R)=1$, for the real line $\R$ with the standard metric.
\label{ex:R}
\end{xmpl}

In the context of statistical learning theory, Nagata dimension captures in an abstract setting the geometry behind Stone's lemma (\cite{stone:77}, Proposition 10) used in the proof of the universal consistency of the $k$-NN classifier in the finite-dimensional Euclidean space (see Sect. 5.3 in \cite{DGL}).

\begin{rmrk}
  In $\R^2=\C$ the family of closed balls of radius one centred at the vectors $\exp(2\pi ki/5)$, $k=1,2,\ldots,5$, has multiplicity $5$ (they all contain the origin) and is disconnected, that is, none of the balls contains the centre of another. Therefore, by Proposition \ref{ex:famdesconexa}, the Nagata dimension of $\ell^2(2)$ is at least $4$.

  On the other hand, suppose, towards a contradiction, that six closed balls $\bar B_{r_i}(x_i)$, $i=1,2,\ldots,6$ form a disconnected family and contain a common point, $x$. It follows that $x$ is different from all the centres. For some $i\neq j$ we have $\angle(x_i-x,x_j-x)\leq \pi/3$. Suppose for the sake of clarity that $\norm{x-x_i}\leq\norm{x-x_j}$. Now a simple planimetric argument, due to Stone (\cite{stone:77}, Prop. 10; Sect. 5.3 in \cite{DGL}) shows that $\norm{x_i-x_j}\leq\norm{x-x_j}$.
  \begin{figure}[ht]
   \begin{center}
    \scalebox{0.12}{\includegraphics{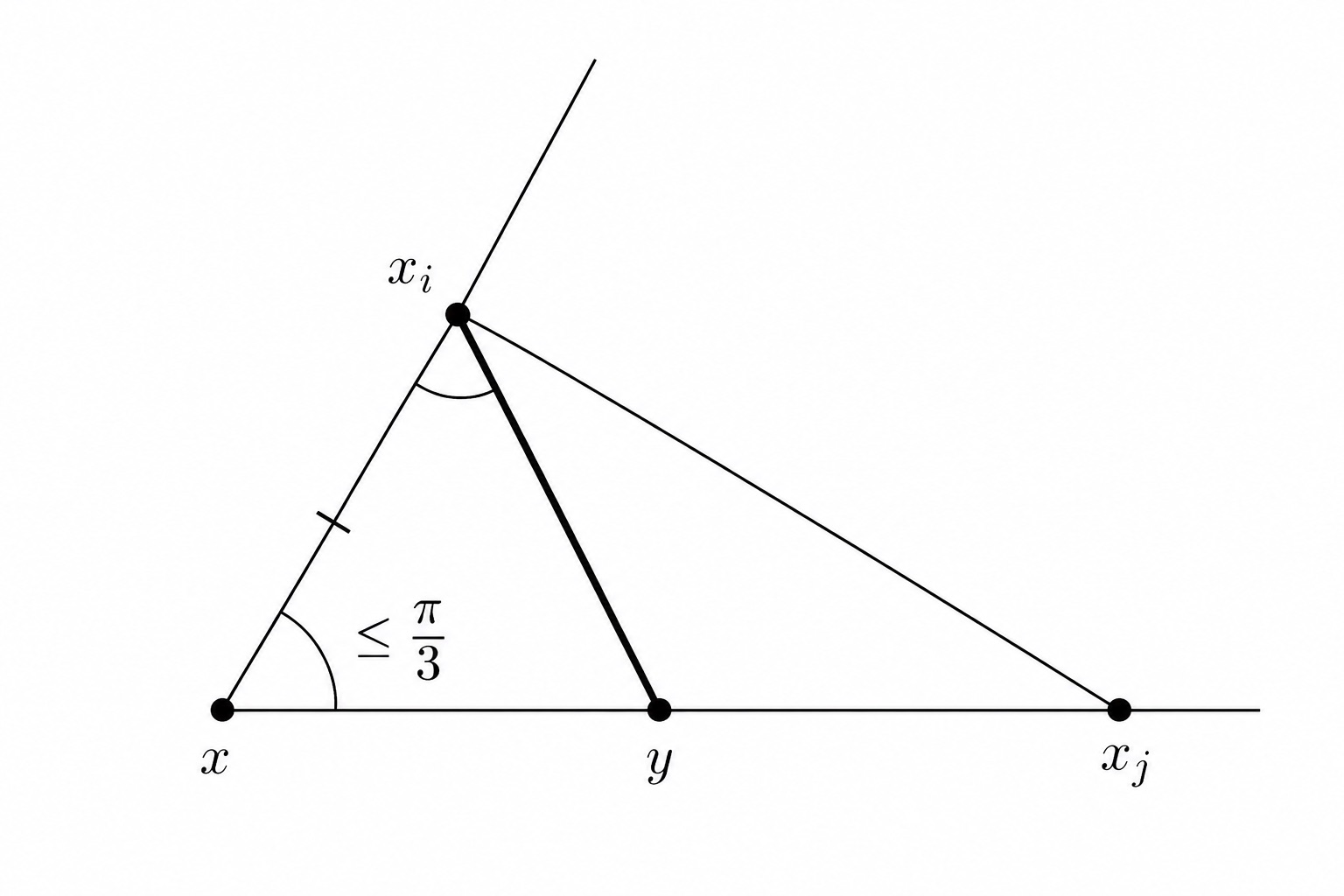}} 
  \end{center}
\caption{Stone's geometric argument.}
  \end{figure}
  
  \begin{align*}
    \norm{x_i-x_j}&\leq \norm{x_j-y}+\norm{x_i-y}\\ & =\norm{x_j-y}+2\norm{y-x}\sin(\angle(y-x,x_i-x)/2)\\
    & \leq\norm{x_j-y}+2\norm{y-x}\sin(\pi/6) \\
    &=\norm{x_j-y}+\norm{y-x} \\
    &= \norm{y_j-x}.
  \end{align*}
  Hence $x_i\in\bar B_{r_j}(x_j)$ and our family of balls is not in fact disconnected, a contradiction. Thus, any disconnected family of closed balls has multiplicity $\leq 5$.
  
  We conclude: $\dim_{Nag}\R^2=4$. The claim that the Nagata dimension of $\R^2$ is $5$, made by Nagata (\cite{nagata_open}, middle of p. 9) without a proof and then repeated by this author \cite{CKP} (Remark 5.7, with a deficient argument), appears inexact. This error was pointed out by the anonymous referee.
  \label{r:R2}
\end{rmrk}

\begin{xmpl}
  The same argument shows that the Nagata dimension of the Euclidean space $\ell^2(d)$ is finite, and it is in particular bounded by $\gamma_d-1$, where $\gamma_d$ is the value of the constant appearing in Stone's geometric lemma as presented in Sect. 5.3 in \cite{DGL}. At the same time, the exact values of the Nagata dimension of the higher-dimensional Euclidean spaces seem to be still unknown, this was stated as an open problem in \cite{nagata_open}. The argument in Remark \ref{r:R2} shows that $\dim_{Nag}(\ell^2(d))$ is exactly the maximal possible number of points on the Euclidean sphere $\mathbb{S}^{d-1}$ that are at a geodesic distance $>\pi/3$ from each other, minus one. But it appears the exact value of this (a bit unusual) variation of the packing number is unknown in a closed form.
%% Indeed, let $x_1,\ldots,x_{\gamma_d+1}$ be points belonging to a ball with centre $x$. Using the argument in the proof of Stone's geometric lemma with $k=1$, mark $\leq \gamma_d$ points $x_i$ belonging to the $\leq \gamma_d$ cones with apex at $x$. At least one point, say $x_j$, has not been marked; it belongs to some cone, which therefore already contains a marked point, say $x_i$, different from $x_j$, and $\norm{x_i-x_j}\leq\norm{x_j-x}$.
\label{ex:cones}
\end{xmpl}

 \begin{rmrk}
A good reference to a great variety of distance dimensions, including the Nagata dimension, even in the absence of triangle inequality, and their applications to measure differentiation, is the article \cite{AQ}.
 \end{rmrk}

 \section{Proof of Theorem \ref{th:main}\label{s:measure}}
 
\subsection{The construction of the learning problem $(\mu,\eta)$}

\subsubsection{The indexing tree}
Let $\Omega$ be a separable complete metric space that is not sigma-finite dimensional in the sense of Nagata. Fix a non-empty closed subspace $\Upsilon\subseteq \Omega$ with the property that none of its non-empty relatively open subsets has a finite Nagata dimension in $\Omega$ (Proposition \ref{p:yx}).

We will work with an unbounded sequence $(m_i)_{i=0}^{\infty}$ of positive natural numbers starting with $m_0=1$, which is arbitrary at this stage, but will be recursively specified at a later point. The only condition we impose from the start is that $m_i\geq 2$ for $i\geq 1$.
Form a rooted prefix tree, $T$, with
vertices being finite words $t=t_1t_2\ldots t_i$, $i\in\N$, $t_j\in [m_j]=\{1,2,\ldots,m_j\}$, $j=1,2,\ldots,i$. For such a $t$, we denote $i=\abs t$.  The root is the empty word, denoted $\ast$; so, $\abs\ast=0$. In this way, the children of the node $t$ with $\abs t=i$ are the words $t1,t2,\ldots,tm_{i+1}$. (Here $t1$ means the word obtained by concatenating the word $t$ with the letter $1$, etc.)

\subsubsection{Selecting points $x^t,y^t$}
\label{ss:properties}
We will now select two families of points of the metric space $\Omega$ indexed with the vertices of the tree $T$, namely, $x^t\in \Omega$ for all $t\in T$,
and $y^t\in \Upsilon$ for $t\in T\setminus\{\ast\}$, as well as numbers $0<\e^t<2^{-\abs t}$, $t\in T$, with the following properties for each $t\in T$, $\abs t =i$, and every $j=1,2,\ldots,m_{i+1}$,  $\ell=1,2,\ldots,m_{i+2}$:

\begin{enumerate}
\item for every $x\in \bar B_{\e^t}(y^{tj})$, the point $x^t$ is strictly closer to $x$ than any point contained in any of the balls $\bar B_{\e^t}(y^{ts})$, $s\neq j$.
  %% , and in any of the balls $\bar B_{\e^{t^\prime}}(y^{t^\prime s})$, where $\abs{t^\prime}=i$, $t^\prime\neq t$ and $s\in[m_{i+1}]$.
\item
  $x^{tj}\in B_{\e^t/3}(y^{tj})$, 
\item $B_{\e^{tj}}(y^{tj\ell})\subseteq B_{\e^t/3}(y^{tj})$,
  %%  \item $\e^{tj}\leq\e^t/3$.
  \item the ball $B_{\e^{t-}}(y^{t})$ does not contain any point of the form $x^{t^\prime}$, where $\abs{t^\prime}\leq i$, other than $x^t$.
  \end{enumerate}

The existence of such points is a direct consequence of our assumptions on $\Omega$ and $\Upsilon$, and we will do it in some detail. Before doing that, let us deduce from (1)-(3) the following property, formally strengthening (1), to be used later on (Subs. \ref{ss:nonat}).

\begin{enumerate}
\item[(5)] Given $t$ with $\abs t =i$, $j\in[m_{i+1}]$, and $x\in \bar B_{\e^t}(y^{tj})$, the point $x^t$ is strictly closer to $x$ than any point contained in any of the balls $\bar B_{\e^{t^\prime}}(y^{t^\prime s})$, where $\abs{t^\prime}=i$, $s\in[m_{i+1}]$, and $t^\prime s\neq tj$.
%%  $t^\prime\neq t$ and $s\in[m_{i+1}]$.
%%   set $r=d(x,x^t)$. The closed ball $\bar B_r(x)$ %%contains no other points $x^{t^\prime}$ with $\abs{t^\prime}\leq i$ except $x^t$, and
%% is disjoint from all the balls of the form $\bar B_{\e^{t^\prime}}(y^{t^\prime s})$, where $\abs{t^\prime}=i$, $s\in[m_{i+1}]$, and $t^\prime\neq t$ or $s\neq j$.
\end{enumerate}

The condition (1) already takes care of the case $t^\prime=t$ and $s\neq j$, so we can assume $t^\prime\neq t$. For every $s\in[m_{i+1}]$
we have $\bar B_{\e^{t^\prime}}(y^{t^\prime s})\subseteq B_{\e^{t^\prime-}}(y^{t^\prime})$ by (3). Again by (1), the balls $\bar B_{\e^{t^\prime-}}(y^{t^\prime})$ and $\bar B_{\e^{t-}}(y^t)$ are disjoint. It remains to show that the ball $\bar B_r(x)$ is contained in the ball $\bar B_{\e^{t-}}(y^t)$.
One has for every $z\in \bar B_r(x)$,
\begin{align*}
  d(z,y^t) &\leq d(z,x)+d(x,y^{t}) \\
  &\leq d(x,x^t) + d(x,y^{t}) \\
  & \leq d(x,y^{t})+d(x^t,y^t)+ d(x,y^{t})\\
\mbox{\tiny by (3) and (2)}  & < \e^{t-},
\end{align*}
meaning $z\in B_{\e^{t-}}(y^t)$.

\subsubsection{Base of recursion}
At the first step, we use the fact that $\Upsilon$ itself is not Nagata finite dimensional in $\Omega$. In particular, as $\dim^s_{Nag}(\Upsilon,\Omega)> m_1$ on each scale $s>0$,
there exists a disconnected collection of $m_1$ balls of radii $<s$ centred in $\Upsilon$ (which balls we can take open by Proposition \ref{p:clopen}), having a point in $\Omega$ in common. At the base step, we take $s=1$.

Denote the centres of the selected balls $y^{1},y^2,\ldots,y^{m_1}\in \Upsilon$, the radii $0<r_1^1,\ldots,r_1^{m_1}<1$, and the common point of the open balls $x^{\ast}\in \Omega$. Since $m_1\geq 2$, it follows that
%% we can assume that
$x^{\ast}$ is distinct from all the centres.
%% by selecting one ball more than necessary and noticing that at most one among them can have $x^\ast$ as the centre.
Take $\e^{\ast}>0$ so small that $\e^{\ast}<1$ and for each $i,j=1,2,\ldots,m_1$, $i\neq j$, we have
\[\e^{\ast} < \frac 14\left(d(y^i,y^j)-d(y^i,x^{\ast})\right).\]
It follows now by the triangle inequality that for any point, $z$, in the closed ball $\bar B_{\e^\ast}(y^i)$ the point $x^{\ast}$ is strictly closer to $z$ than is any point, $w$, of any closed ball $\bar B_{\e^\ast}(y^j)$, $j\neq i$:
\begin{align*}
  d(z,w)-d(z,x^{\ast})  & \geq d(y^i,y^j)- d(y^i,x^{\ast}) -4\e^{\ast} \\
  & > 0.
\end{align*}
By reducing $\e^{\ast}$ further if necessary, we can assume that none of the balls $\bar B_{\e^\ast}(y^i)$ contains the point $x^{\ast}$, to satisfy (4).

\subsubsection{Step of recursion}
Let $i\geq 1$ and suppose that $x^t\in\Omega$ and $\e^t>0$ have been selected for all $t$ with $0\leq\abs t<i$, and $y^t\in\Upsilon$ have been chosen for all $t$ with $1\leq\abs t\leq i$, in a way to satisfy the properties (1)-(3) in subs. \ref{ss:properties}.

Fix a $t$ with $\abs t=i$ and denote $t^-$ the $(i-1)$-prefix of $t$, that is, $t$ with the last letter removed.
We now have to select $x^t\in \Omega$ and $y^{tj}\in \Upsilon$, $j=1,2,\ldots,m_{i+1}$, all within the open ball $B_{\e^{t^-}/3}(y^{t})$, as well as a value $0<\e^t<2^{-i}$.

The relative open ball $B_{\e^{t^-}/6}(y^{t})\cap \Upsilon$ is non-empty (it contains $y^t$) and so is not finite dimensional in $\Omega$ in the sense of Nagata by our choice of $\Upsilon$, meaning it is infinite dimensional in $\Omega$ on all scales, including the scale $s=\e^{t^-}/6$. There exist points $y^{t1},y^{t2},\ldots,y^{tm_{i+1}}\in B_{\e^{t^-}/6}(y^t)\cap \Upsilon$ and radii $0<r^{t1},r^{t2},\ldots,r^{tm_{i+1}}<\e^{t^-}/6$ such that the open balls $B_{r^{t1}}(y^{t1}),B_{r^{t2}}(y^{t2}),\ldots,B_{r^{tm_{i+1}}}(y^{tm_{i+1}})$ are disconnected (do not contain each other's centres) and have a point in $\Omega$ in common, which point we denote $x^t$. As $m_{i+1}>1$, we see 
%% by selecting an excessive number of balls and then discarding one as necessary, we can assure
that all the centres $y^{ti}$ are different from the point $x^t$.
%%as well as from the (finitely many) points of the form $x^{t^\prime}$ as previously selected.

By the triangle inequality, all open balls $B_{r^{tj}}(y^{tj})$, $j\in [m_{i+1}]$ are contained in $B_{\e^{t^-}/3}(y^t)$. In particular, $x^t\in B_{\e^{t^-}/3}(y^t)$. Just like at the first step of recursion, we can find $\e^t\in (0,2^{-i})$ with the property that for every $j\in[m_{i+1}]$ and every point $z\in \bar B_{\e^t}(y^{tj})$, the point $x^t$ is strictly closer to $z$ that any point of any closed ball $\bar B_{\e^t}(y^{ts})$, $s\neq j$. By its choice, $\e^t$ is strictly less than any $r^{tj}$, and so the condition (3) is satisfied too. Further reducing $\e^t$ if need be, we can assume that none of the balls $\bar B_{\e^t}(y^{tj})$ contains $x^t$. Jointly with the conditions (1)-(3), this implies the formally stronger condition (4): first one sees, inductively, that the points $x^{t^\prime}$, $\abs{t^\prime}\leq i$, with $t^{\prime}\neq t$, cannot belong to $B_{r^{tj}}(y^{tj})$, and second, the balls $B_{r^{tj}}(y^{tj})$ for different $j$ are pairwise disjoint by (1), and in each one of them we select exactly one point of the form $x^{tj}$.  
%% or any points of the form $x^{t^\prime}$ as previously selected by our recursive procedure.
The recursion step is now completed.

\subsubsection{Compact set $K$\label{s:K}}
By the condition \ref{ss:properties}(3), for any $t$ with $\abs t=i\geq 1$ and any word $s$, we get
$d(y^t,y^{ts})<2^{-i+1}$. Also, the condition \ref{ss:properties}(2) says that $d(y^t,x^t)<2^{-i+1}$. Consequently, all the points $y^t$, $x^t$ with $\abs t\geq i$ are contained within the distance $2^{-i+2}$ from the finite set of points $y^t$, $\abs t=i$. This implies the entire countable set of points $x^t$, $\abs t\geq 0$, $y^t$, $\abs t\geq 1$ is precompact. Denote the closure of this set $K$. It is a compact subset of the complete metric space $\Omega$. The above mentioned property of $K$ implies that $K$ can be written as the union of the set $\{x^t\colon t\in T\}$ and the closure of the set $\{y^t\colon t\in T\setminus\{\ast\}\}$.

We will now construct two finite Borel measures whose support is contained in $K$, each one of total mass $1/2$. The measure $\mu_1$ is purely atomic and has $x^t$, $t\in T$ as atoms. The non-atomic measure $\mu_0$ is a weak limit of the sequence of uniform measures supported on the points $y^t$, $\abs t=i$, as $i\to\infty$. 

\subsubsection{Measure $\mu_1$} Fix a sequence $(\gamma_n)_{n=0}^{\infty}$, $\gamma_n>0$, with $\sum_{n=0}^\infty\gamma_n=1/2$. Now define for each $t\in T$
\begin{equation}
  \mu_1\{x^t\} = \frac{\gamma_{\abs t}}{\prod_{0\leq j\leq\abs t}m_j}.
  \label{massmu_1}
\end{equation}
  Notice that for each $i=0,1,\ldots$,
  \[\left\vert \{x^t\colon \abs t = i\}\right\vert = \prod_{0\leq j\leq i}m_j,\]
  and so $\mu_1\{x^t\colon \abs t =i\}=\gamma_i$, and
  $\sum_{t\in T}\mu_1\{x^t\}=\mu_1(K) = 1/2$.

  \subsubsection{Measure $\mu_0$\label{ss:mu0}}
  For every $i=1,2,\ldots$ let $\mu_0^{(i)}$ be a uniform measure of total mass $1/2$ supported on the points $y^t$ with $\abs t =i$, that is,
\[\mu^{(i)}\{y^t\} = \frac{1}{2\prod_{1\leq j\leq i}m_j}.\]
Since for each  $i=1,2,\ldots$,
\[\left\vert \{y^t\colon \abs t = i\}\right\vert = \prod_{1\leq j\leq i}m_j,\]
we have $\mu^{(i)}_0(K)=1/2$ for each $i$.

The sequence of measures $(\mu_0^{(i)})_{i=1}^{\infty}$ converges in the weak topology in the space of finite measures on the compact space $K$.
Indeed, the weak topology (that is, the weak$^\ast$ topology on the dual space of $C(K)$) on the measures of fixed mass on a compact space is metrized by the mass transportation distance (see e.g. \cite{V}, corollary 6.13). For every $t$, $\abs t\geq 1$, the $\mu_0^{(i)}$-measure of the ball $\bar B_{\e^{t^-}}(y^t)$ is equal to $1/(2\prod_{j\leq i}m_i)$, because the ball only contains exactly one point among all $y^t$ with $\abs t=i$, namely, its centre. If now $j>i$, then the number of points $y^s$ with $\abs s =j$ contained in the same ball $\bar B_{\e^{t^-}}(y^t)$ is equal to $\prod_{\ell=i+1}^{j}m_\ell$, and so the total mass of the ball under the measure $\mu_0^{(j)}$ is the same as under $\mu_0^{(i)}$. The transport of mass between the restrictions of the two measures on the ball requires shifting one of them by a distance $\leq \e^{t^-}<2^{-i}$. We conclude that the overall mass transportation distance between $\mu_0^{(i)}$ and $\mu_0^{(j)}$ is bounded by $2^{-i}$ times the total mass, $1/2$.
This way, the sequence $(\mu_0^{(i)})$ is Cauchy, hence convergent in the weak topology on the compact space of measures of mass $1/2$. Denote the limit $\mu_0$. It also satisfies $\mu_0(K)=1/2$.

For every $i$, the union of the closed balls  $\bar B_{\e^{t^-}}(y^t)$, $\abs t=i$ has full $\mu_0$-measure, $1/2$. In particular, it follows that
the measure $\mu_0$ is diffuse, because for each $t$ with $\abs t=i$, it has the property $\mu_0(\bar B_{\e^{t^-}}(y^t))=1/(2\prod_{0\leq j\leq i}m_i)\to 0$ as $i\to\infty$. (Remembering that $m_i\geq 2$.)

\subsubsection{The probability measure $\mu$ and the regression function $\eta$}
Since $\mu_1$ is purely atomic and $\mu_0$ is diffuse, the two measures are mutually singular. 
Define a probability measure $\mu=\mu_0+\mu_1$. Since  $\mu_0\perp\mu_1$, there is a Borel function $\eta\colon X\to \{0,1\}$ that is $\mu_0$-a.e. equal to $0$ e $\mu_1$-a.e. equal to $1$. (The indicator function of the set $\{x^t\colon t\in T\}$ is such.)

This $(\mu,\eta)$ will become our learning problem. The Bayes (optimal) classifier for this problem is a.e. equal to the binary function $\eta$, so
assumes the values $0$ and $1$ with probability half each. 
As we will see, given any sequence of values $(k_n)_{n=1}^{\infty}$ of values of $k$ with $k_n/n\to 0$, under a suitable selection of the sequence $(m_i)$ used to construct $\mu$ and $\eta$,
the $k$-NN rule with $k=k_n$ does not asymptotically converge to $\eta$ in probability.

Namely, for a certain subsequence $(n_i)$ of the values of $n$, the sequence of the predictions of the $k$-NN classifier under $(\mu,\eta)$, corresponding to the values $n=n_i$ and $k=k_{n_i}$, will converge in probability, as $i\to\infty$, to the constant predictor $1$. This, the $k$-NN classifier will not be consistent under this problem.

\subsection{Choosing the sequences $(m_i)$ and $(n_i)$}
Let a sequence $(k_n)_{n=1}^{\infty}$ of values of $k$ be given, satisfying $k_n/n\to 0$. (We do not require $k_n\to\infty$.)
Now we will select the sequences $(m_i)$ and $(n_i)$.

Fix a summable sequence of values of the risk $\delta_i>0, i\in\N$.
Remember that $m_0=1$. This is the base of the recursion.

\subsubsection{Step $m_i\to n_i$\label{s:k}}
Let $i=0,1,\ldots$, and suppose that the values $m_j$, $j=1,2,\ldots,i$ have already been selected. We want to choose $n_i$ so large that, with confidence $1-\delta_i$, each of the points $x^t$, $\abs t=i$ appears among the i.i.d. random elements $X_1,\ldots,X_{n_i}$, following the law $\mu$ as above, no fewer than $k_{n_i}$ times. Remembering that the mass of $x^t$ is given by eq. (\ref{massmu_1}) and that $k_n/n\to 0$, first assume that $n$ is so large that
\begin{equation}\frac{k_n}{n}<\frac 12\mu\{x^t\}=\frac 12\mu_1\{x^t\}=\frac{\gamma_{i}}{2\prod_{0\leq j\leq i}m_j}.
  \label{eq:kn}
\end{equation}
By the Chernoff bound, the probability for the empirical measure of the singleton $\{x^t\}$ to be smaller than half its expectation, $\mu\{x^t\}$, is limited by
\[\exp\left[{-2n\left(\frac{\gamma_{i}}{2\prod_{0\leq  j\leq i}m_i}\right)^2}\right].\]
The probability that such a deviation happens for at least one among the points $x^t$, $\abs t=i$, is limited by the union bound by
\[\prod_{0\leq j\leq i}m_j\exp\left[{-2n\left(\frac{\gamma_{i}}{2\prod_{0\leq j\leq i}m_j}\right)^2}\right].\]
We want the above to be limited by $\delta_i$. Taking the natural logarithm of the inequality,
we arrive at
\[ \frac{\gamma^2_{i}n}{2\prod_{0\leq j\leq i}m^2_j}> \sum_{0\leq j\leq i}\log m_j-\log\delta_i,\]
that is,
\[n>\frac{2\prod_{0\leq j\leq i}m^2_j}{\gamma^2_{i}} \left(\sum_{0\leq j\leq i}\log m_j-\log\delta_i \right).\]
Choose any value $n=n_i$ satisfying the above inequality and such that eq. (\ref{eq:kn}) holds. If now $n= n_i$, then, with confidence $1-\delta_i$, the empirical measure of every point $x^t$, $\abs t=i$ supported on a random $n_i$-sample will be at least $(1/2)\mu\{x^t\}>k_{n_i}/n_i$. In other words, with confidence $1-\delta_i$, each point $x^t$, $\abs t=i$ will
appear at least $k_{n_i}$ times among the members of a random i.i.d. $n_i$-sample following the law $\mu$.

\subsubsection{Step $n_i\to m_{i+1}$}
\label{ss:fewer}
Now we will choose $m_{i+1}$. We want this number to be so large that no more than a fraction of $\delta_i$ among the balls of the form $B_{\e^t}(y^{tj})$, $\abs t =i$, $j=1,2,\ldots,m_{i+1}$ contain at least $k_{n_i}/2$ elements of a random i.i.d. $n_i$-sample following the law $\mu$. The number of balls containing $\geq k_{n_i}/2$ elements of a $n_i$-sample is limited by $2n_i/k_{n_i}$. So it is enough to require
\[\frac{2n_i}{k_{n_i}} < \delta_i \prod_{j=0}^{i+1}m_j = \delta_i \left(\prod_{j=0}^{i}m_j\right) m_{i+1},\]
that is,
\[m_{i+1}> \frac{2n_i}{k_{n_i}\delta_i\prod_{j=0}^{i}m_j}.\]
Now the recursive step is accomplished.

Notice in passing that as $n_i/\prod_{j=0}^{i}m_j\geq k_{n_i}$ by eq. (\ref{eq:kn}), the r.h.s. of the above expression is no less than $2/\delta_i$, and so $m_i\to\infty$. (This assures that the measure $\mu_0$ is diffuse, see the end of Subsection \ref{ss:mu0}, though for that it would already suffice that each $m_i\geq 2$.)

\subsection{Absence of consistency of the $k$-NN rule under $(\mu,\eta)$}

\subsubsection{Predictor for atoms\label{ss:xt}}
Each point $x^t$ is an atom, labelled deterministically $1$. It is well known and easily proved that on the set of such atoms the $k$-NN classifier in the limit $n\to\infty$ will converge in probability to the identical value $1$, as long as $k/n\to 0$.

It remains to analyse the situation with $x\in K$ that is not of the form $x^t$ and thus will $\mu$-almost surely be labelled $0$. We will show that on the set of such points, the label predicted by the $k$-NN classifier does {\em not} converge in probability to zero.

\subsubsection{Predictor for non-atomic points\label{ss:nonat}}
Fix $i\in\N$, $i\geq 1$ throughout the subsection.

Going back to the definition of the compact set $K$ (Subsection \ref{s:K}), we notice that, since each point $y^t$ has zero measure, $\mu$-almost all elements $x$ of the non-atomic part of $K$, that is, of $K\setminus\{x^t\colon t\in T\}$ (in fact, all except possibly a finite number) are contained in the set
$\cup_{\abs t=i}\cup_{j=1}^{m_{i+1}}B_{\e^t}(y^{tj})$. Since for $i$ fixed the latter balls are pairwise disjoint, for $\mu$-almost every non-atomic $x$ there is a unique $t\in T$ with $\abs t =i$ and $j=1,2,\ldots,m_{i+1}$ such that $x\in B_{\e^t}(y^{tj})$.

Form a closed ball $\bar B_r(x)$ around $x$ of radius $r=d(x,x^t)$. 
By the property \ref{ss:properties}(5), its intersection with $\cup_{\abs t=i}\cup_{j=1}^{m_{i+1}}B_{\e^t}(y^{tj})$ is the same as the intersection with $B_{\e^t}(y^{tj})$ alone. Also, $\bar B_r(x)$ contains the point $x^t$, and, by the property \ref{ss:properties}(4), cointais no other points of the form $x^{t^\prime}$, $\abs{t^\prime}\leq i$.
%% Indeed, 
%% %% $t^\prime\neq t$:
%% for $\abs{t^\prime}=i$, this follows from \ref{ss:properties}(1) and \ref{ss:properties}(4), and for $\abs{t^\prime}<i$, again from \ref{ss:properties}(4),
%% %% as
%% $\bar B_r(x)\subseteq B_{\e^{t-}}(y^t)$, and the latter ball contains no  
%% such points except $x^t$ by property \ref{ss:properties}(4).

%% Now, $x^t$ is strictly closer to $x$ than any point of any other ball of the same form, $B_{\e^t}(y^{ts})$, $s\neq j$ (property \ref{ss:properties}(1)). The property \ref{ss:properties}(3) says that the union of the above balls is contained, in its turn, in the ball $B_{\e^{t-}}(y^{t})$. We have $d(x,x^t)<\e^t<2^{-i}$, so 
%% If we form a closed ball around $x$ of radius $d(x,x^t)$, the intersection of this ball with $K$ will be the same as the intersection of the ball $B_{\e^t}(y^{tj})$ with $K$ plus $\{x^t\}$.

Hence, with probability $>1-\delta_i$, conditionally on $X\notin \{x^t\colon t\in T\}$, the {\em open} ball around a random element $X\sim\mu$ having the same radius as the distance to the nearest $x^t$, $\abs t \leq i$, contains strictly fewer than $k_{n_i}/2$ elements of the random i.i.d. $n_i$-sample following the law $\mu$ (Subsection \ref{ss:fewer}). This means that the radius of the smallest {\em closed} ball containing at least $k_{n_i}$ nearest neighbours of $X$ has to be greater than or equal to the distance $d(X,x_t)$.

At the same time, 
with probability $>1-\delta_i$, each of the points $x^t$, $\abs t=i$ appears at least $k_{n_i}$ times among the elements of the random i.i.d. $n_i$-sample following the law $\mu$ (Subs. \ref{s:k}). Jointly the two properties mean that, with probability $>1-2\delta_i$, the smallest closed ball around a random element $X$ containing at least $k_{n_i}$ nearest neighbours of $X$ is exactly the distance from $X$ to the nearest point of the form $x^t$. The ``missing part'' of the $k_{n_i}$ nearest neighbours to $X$ will all be equal to the point $x^t$, all labelled $1$.

We conclude:  with probability $>1-2\delta_i$, strictly more than half of the $k=k_{n_i}$ nearest neighbours to $X$ among the random i.i.d. $n_i$-sample are equal to one of the points $x^t$, $\abs t=i$, namely the one nearest to $X$. And the points $x^t$ are almost surely labelled $1$.

This means that, with probability $>1-2\delta_i$, a random element $X\in K\setminus \{x^t\colon t\in T\}$ will get labelled $1$ by the $k$-NN classifier at the step $(n_i,k_{n_i})$.

\subsubsection{$i\to\infty$}
Combining the observations made in subsections \ref{ss:xt} and \ref{ss:nonat}, we conclude that, as $\sum\delta_i<\infty$,
%% by Borel--Cantelli,
the subsequence of $k$-NN classifiers corresponding to $(n_i,k_{n_i})$ converges in probability to the identical value $1$ both over $\Omega\setminus \{x^t\colon t\in T\}$ and over $\{x^t\colon t\in T\}$.
%% almost surely, in particular,
%%in probability. Combining this with the observation in Subs. \ref{ss:xt}, we conclude:
Thus, the subsequence of $k$-NN classifiers corresponding to the values $n=n_i$, $k= k_{n_i}$ converges to the identical value $1$ over all of $\Omega$ in probability as $i\to\infty$.

This implies that the sequence of $k$-NN classifiers corresponding to the original sequence $(n,k_n)$ does {\em not} converge in probability to the Bayes classifier, $\eta$, and so the $k$-NN rule is not consistent under $(\mu,\eta)$.

\section{The weak Lebesgue--Besicovitch property is insufficient for the $k$-NN classifier consistency\label{s:weak}}

Here is another corollary obtained by combining our main result (Theorem \ref{th:main}) with known constructions from real analysis in metric spaces.

\begin{crllr}
  There exists a complete separable metric space equipped with a Borel probability measure $\mu$, satisfying the weak Lebesgue--Besicovitch property for every $L^1(\mu)$-function, yet in which the $k$-NN classification rule is not universally consistent for any sequence of values of $k_n\to\infty$, $k_n/n\to 0$.
  \label{c:notimplies}
  \end{crllr}

A source of metric spaces in which every Borel probability measure satisfies the {\em weak} Lebesgue--Besicovitch property is given by the following concept, which, just like the Nagata dimension, has originated in the topological dimension theory, but is weaker (less restrictive).

\begin{dfntn}[\cite{dG}; \cite{AQ}, 3.5, 3.7]
Let $\delta\in\N$.
A metric space $\Omega$ has de Groot dimension $\leq\delta$ (on the scale $s=+\infty$) if every finite family of closed balls having the same radii admits a subfamily covering all the centres of the original balls and having multiplicity $\leq\delta+1$. The de Groot dimension of $\Omega$ is the smallest $\delta$ with this property, if it exists, and $+\infty$ otherwise. We denote it $\dim_{dG}(\Omega)$.
\end{dfntn}

Comparing this concept with Definition \ref{d:dimnagata}, we conclude that de Groot dimension is bounded by the Nagata dimension: $\dim_{dG}(\Omega)\leq\dim_{Nag}(\Omega)$ (on the scale $+\infty$, the only one we will need). The distinguishing examples between the two dimensions are easy to construct. For instance, the convergent sequence $2^{-n}e_n$, $n=1,2,\ldots$ together with the limit $0$ in the Hilbert space $\ell^2$ (where $e_n$ are standard orthonormal basic vectors) has de Groot dimension $2$ yet its Nagata dimension is infinite (see \cite{KP}, Example 3.12).

\begin{thrm}[Assouad and Quentin de Gromard, \cite{AQ}, Prop. 3.3.1(b)+Prop. 3.1]
Let a complete separable metric space $\Omega$ have finite de Groot dimension. Then for every locally finite Borel measure $\mu$ on $\Omega$, every $L^1(\mu)$-function $f\colon\Omega\to\R$ satisfies the weak Lebesgue--Besicovitch differentiation property: 
\begin{equation}
\frac{1}{\mu(B_r(x))}\int_{B_r(x)} f(x^\prime)\,d\mu(x^\prime) \to f(x)~~\mbox{when $r\to 0$}
\label{eq:lbg}
\end{equation}
in measure.
\end{thrm}

Thus, in order to establish Corollary \ref{c:notimplies}, it is enough, in view of Theorem \ref{th:main}, to give examples of complete separable metric spaces having finite de Groot dimension, which are at the same time not sigma-finite dimensional in the sense of Nagata. 

While it may not be so easy to verify the finiteness of de Groot dimension directly, in view of its combinatorial flavour, the following is an important subclass of metric spaces having finite de Groot dimension.
Recall that a metric space $(X,d)$ has the {\em doubling property} if there is a constant $C\geq 1$ such that for every $r>0$, every closed ball of radius $2r$ is covered by $\leq C$ closed balls of radii $r$.

\begin{prpstn}
  A metric space $\Omega$ having the doubling property with constant $C$ has finite de Groot dimension with $\dim_{dG}(\Omega)\leq C-1$.
  \label{p:doubling}
  \end{prpstn}

\begin{proof}
  Let $\bar B_r(x_i)$, $i=1,2,\ldots,n$ be a finite family of closed balls all having the same radius $r>0$. Select a maximal set of indices, $J\subseteq [n]$, such that the centres $x_i,x_j$ for $i,j\in J$, $i\neq j$ are at a distance $>r$ between themselves. Clearly, the balls $\bar B_r(x_i)$, $i\in J$ still contain all the original centres, by the maximality of $J$. 
  Let now $x\in\Omega$ be any. Cover the ball $\bar B_r(x)$ with balls $\bar B_{r/2}(y_k)$, $k=1,2,\ldots,C$. Each ball of radius $r/2$ can only contain at most one element $x_i$, $i\in J$. This means no more than $C$ balls of the form $\bar B_r(x_i)$, $i\in J$ contain $x$, and so the multiplicity of the subfamily $\bar B_r(x_i)$, $i\in J$ is $\leq C$. 
  \end{proof}

\begin{xmpl}[K\"aenm\"aki, Rajala and Suomala, Example 5.6 in \cite{KRS}] There exists a metric $\rho$ on the Cantor space $C$ (realized as a product of a sequence of finite spaces) having the doubling property and yet such that with regard to $\rho$ and a suitable product probability measure on $C$ the strong Lebesgue--Besicovitch differentiation property fails.

  In view of the theorem of Preiss--Assouad--Quentin de Gromard (Subsection \ref{subs:pag}), the metric space $\left(C,\rho\right)$ is not sigma-finite dimensional in the sense of Nagata. At the same time, by Proposition \ref{p:doubling}, this metric space has finite de Groot dimension. This, in view of Theorem \ref{th:main}, gives perhaps the simplest example with a desired combination of properties to validade Corollary \ref{c:notimplies}.
  \label{ex:krs}
\end{xmpl}

The second example is classical.

\begin{xmpl}[Heisenberg group] 
The Heisenberg group $\Hg$ is the Euclidean space $\R^{3}$ equipped with the following group multiplication:
\begin{equation}
(x,y,z)\cdot (x^{\prime},y^{\prime},z^{\prime}) = 
\left(x+x^{\prime},y+y^{\prime},z+z^{\prime}-2 xy^{\prime}+2 yx^{\prime}\right).
\label{eq:multi1}
\end{equation}
The identity of this group is the zero element $(0,0,0)$ of $\R^3$, and the inverse of $(x,y,z)$ is the same as the usual additive inverse $(-x,-y,-z)$.

The formula 
\[\left\vert (x,y,z) \right\vert_{\Hg} = \left((x^2+y^2)^2+z^2 \right)^{1/4}\]
defines a {\em group norm} on $\Hg$, in the sense that $\abs{p^{-1}}_{\Hg}=\abs{p}_{\Hg}$ and $\abs{p\cdot q}_{\Hg}\leq \abs{p}_{\Hg}+\abs{q}_{\Hg}$.
The latter inequality is not quite trivial to verify, and follows from a more general inequality of Cygan \cite{cygan}. As a consequence,
the formula
\[d(p,q)=\abs{p^{-1}\cdot q}_{\Hg},~~p,q\in\Hg\]
defines a metric on the group $\Hg$, which is easily seen to be left-invariant and compatible with the Euclidean topology. This metric is known as either ({\em Cygan}-){\em Kor\'anyi}, or else
{\em Carnot} distance. See e.g. \cite{leD} for more.

The doubling property for the above metric immediately follows from  the fact it is not only left-invariant, but also {\em homogeneous}: for $t>0$, the transformation $(x,y,z)\mapsto (tx,ty,t^2z)$ increases the distance between any pair of points by the factor of $t$. In this way, it is enough to cover the closed ball of radius one around identity with finitely many balls of radius $1/2$, and this holds because of local compactness. We conclude that the Heisenberg group has finite de Groot dimension.

At the same time, it was shown, independently, by Kor\'anyi and Reimann (\cite{KR}, p. 17) and Sawyer and Wheeden (\cite{SW}, Lemma 4.4, p. 863), that the Nagata dimension of the Heisenberg group $\Hg$ is infinite, and in fact their argument shows more: that $\Hg$ is not sigma-finite dimensional in the sense of Nagata (see \cite{KP}, Corol. 3.20).

Thus, the Heisenberg group gives another way to establish Corollary \ref{c:notimplies} of Theorem \ref{th:main}.
  \end{xmpl}

\begin{rmrk}
 We use this opportunity to correct a wrong statement made in the previous article \cite{KP}, namely, Corollary 3.18, where we erroneously claimed that the $k$-NN classifier is universally consistent in the Heisenberg group. The main result of the present paper, Theorem \ref{th:main}, implies otherwise.
  \end{rmrk}

\section{An equivalent metric on $\R$ under which the $k$-NN classifier fails\label{s:R}}

Recall that two metrics, $d$ and $\rho$, on the same set $X$ are {\em uniformly equivalent} if for every $\e>0$ there is $\delta>0$ so that
\[\forall x,y\in X,~~d(x,y)<\delta\Rightarrow \rho(x,y)<\e,~~\rho(x,y)<\delta\Rightarrow d(x,y)<\e.\]
In particular, such $d$ and $\rho$ are equivalent, that is, induce the same topology.
Two uniformly equivalent metrics share the same Cauchy sequences, so if $X$ is a complete metric space under one of the two uniformly equivalent metrics, it is complete under the other. Any two compatible metrics on a compact space are easily verified to be uniformly equivalent.

\begin{crllr}
  Let $(X,d)$ be a complete separable metric space of uncountable cardinality. There exists a metric $\rho$ on $X$ uniformly equivalent to $d$ and such that the $k$-NN classifier is not universally consistent in the (complete, separable) metric space $(X,\rho)$, under any choice of sequence $k_n\to\infty$, $k_n/n\to 0$.
  \label{c:equiv}
 \end{crllr}

In particular, already the following feels a bit counter-intuitive.

\begin{crllr}
  There is a metric $\rho$ on the real line $\R$, uniformly equivalent to the usual metric, and such that the $k$-NN classifier is not universally consistent in the (complete, separable) metric space $(\R,\rho)$, under any choice of sequence $k_n\to\infty$, $k_n/n\to 0$.
\end{crllr}

We will first collect some results used in the proof.

\begin{xmpl}[Davies \cite{davies}, Theorem II] There exists a compact metric space $\Omega$ of diameter one and two Borel measures $\mu_1,\mu_2$ with $\mu_1(\Omega)=1/3$, $\mu_2(\Omega)=2/3$, such that $\mu_1(B)=\mu_2(B)$ for every closed ball in $\Omega$ of radius less than one.

  The distances in the space $\Omega$ all take values of the form $2^{-i}$ ({\em ibid.}, p. 159$_{-8}$), which implies that $\Omega$ is zero-dimensional and, as by the construction it has no isolated points, is homeomorphic to the Cantor set $\{0,1\}^{\N}$ by the Brouwer theorem (\cite{kechris}, Theorem 7.4).

  Denote $\eta_1,\eta_2$ the Radon-Nikod\'ym derivatives of $\mu_1$ and $\mu_2$, respecively, with regard to the probability measure $\mu_1+\mu_2$. They are obviously different between themselves, so at least one of $\eta_1,\eta_2$ fails the Lebesgue differentiation property.
  
  The theorem of Preiss--Assouad--Quentin de Gromard (see Subsection \ref{subs:pag}) implies that the Davies metric space $\Omega$ is not sigma-finite dimensional.
  \label{ex:davies}
\end{xmpl}

The following is a classical result from the early days of general topology and descriptive set theory.

\begin{thrm}[See e.g. \cite{kechris}, Theorems 6.2 and 6.4]
  Every complete separable metric space of uncountable cardinality contains a homeomorphic copy of the Cantor space.
  \label{th:emb}
\end{thrm}

And here is one of the recent variations on the theorem of Hausdorff about extending metrics from subspaces (see e.g. \cite{torunczyk}).

\begin{thrm}[Nguyen To Nhu, \cite{TN}, Corollary 3.5]
  Let $K$ be a compact subspace of a metric space $(X,d)$, and let $\rho$ be a compatible metric on $K$. Then $\rho$ extends to a metric $\bar\rho$ on $X$ uniformly equivalent to $d$.
  \label{th:extend}
\end{thrm}

\begin{proof}[Proof of Corollary \ref{c:equiv}]
  Let $X=(X,d)$ be a complete separable metric space of uncountable cardinality.
  Identify the Davies space $\Omega\cong\{0,1\}^{\N}$ (Example \ref{ex:davies}) with a topological subspace of $X$ (Theorem \ref{th:emb}), and extend the metric from $\Omega$ to a metric $\rho$ over all of $X$ that is uniformly equivalent to $d$ (Theorem \ref{th:extend}). Since the metric on $\Omega$ is not sigma-finite dimensional, the same is true for its extension $\rho$. According to our Theorem \ref{th:main}, the $k$-NN classifier is not universally consistent in $(X,\rho)$.
\end{proof}

\begin{rmrk}
  Our Corollary \ref{c:equiv} was inspired by the (apparently unpublished) result of Andretta, Camerlo and Costantini (\cite{ACC1}, slide 12): every separable complete metric space equipped with a non-atomic probability measure, $(X,d,\mu)$, admits a complete compatible metric $\rho$ such that $(X,\rho,\mu)$ fails the strong Lebesgue--Besicovitch differentiation property.

  The result uses a strengthened form of Theorem \ref{th:emb}, namely, Theorem 3.1 in \cite{ACC}, affirming that the Cantor set with the canonical Borel probability measure (Haar measure) admits a measure-preserving homeomorphic embedding into any complete separable metric space $X$ equipped with a non-atomic measure $\mu$ if for some Borel set, $A\subseteq X$, one has $1<\mu(A)<\infty$. Another ingredient is the example of K\"aenm\"aki, Rajala and Suomala from \cite{KRS}, mentioned above as Example \ref{ex:krs}, where a metric with a doubling property is constructed on the Cantor set with regard to which the standard Borel probability measure fails the Lebesgue--Besicovitch property. Here the doubling property is not used, so the (much earlier) Davies example could have been used instead.

  Let us remark that if instead of the Hausdorff extension theorem one applies the result of Nguyen To Nhu \cite{TN} (Theorem \ref{th:extend}), one can further strengthen the above result by Andretta, Camerlo and Costantini by assuming the metric $\rho$ to be uniformly equivalent to $d$ as well.

  Of course all we needed in order to apply our Theorem \ref{th:main}, is the existence on $X$ of a uniformly equivalent metric $\rho$ that is not sigma-finite dimensional, so a much simpler argument suffices.
\end{rmrk}

\section{The $1$-nearest neighbour classifier\label{s:1NN}}

Here we aim to prove Theorem \ref{th:1NN}. Let us recall that the result states the equivalence, for every complete metric space $\Omega$, of the following two conditions between themselves and with all the three equivalent conditions in Theorem \ref{th:equivalences}:

   \begin{enumerate}
   \item[(4)] Under every distribution of the labelled data on $\Omega\times\{0,1\}$, the misclassification error of the $1$-nearest neighbour rule, asymptotically as the size of an i.i.d. labelled sample goes to infinity, does not exceed $2\ell^{\ast}$, where $\ell^{\ast}$ is the Bayes error.
     \item[(4$^\prime$)] Under every distribution of the labelled data on $\Omega\times\{0,1\}$ whose Bayes error is zero (that is, the regression function is deterministic), the $1$-nearest neighbour rule is consistent when the size of an i.i.d. labelled sample goes to infinity.
   \end{enumerate}

   The implication (4)$\Rightarrow$(4$^\prime$) is trivially true. The implication (4$^\prime)\Rightarrow$ Theorem \ref{th:equivalences}(3)
   follows from Theorem \ref{th:main}, because the regression function constructed therein is deterministic and the result holds for any sequence $(k_n)$ of natural numbers satisfying $k_n/n\to 0$, including the constant sequence $k_n=1$. Thus, it suffices to prove that the strong Lebesgue--Besicovitch condition (2) in Theorem \ref{th:equivalences} implies the property (4) above. We devote to this the rest of the Section, borrowing techniques from Sections 5.2 and 5.4 in \cite{DGL} and from Section 2.2 in \cite{CG}.

   Let $\Omega$ be a complete separable metric space, equipped with a Borel probability measure $\mu$ and a regression function $\eta$ that satisfied the strong Lebesgue--Besicovitch differentiation property with regard to $\mu$.
Let $(X_1,\ldots,X_n,Y_1,\ldots,Y_n)$ be a random i.i.d. $n$-sample, where $X_i\in\Omega$ are i.i.d. random variables following the law $\mu$ and $Y_i$ are independent Bernoulli random variables with probabilities of success $\eta(X_i)$. To break the distance ties, we use auxiliary i.i.d. real random variables $\xi_1,\ldots,\xi_n$ following some non-atomic law and independent of the remaining random variables.
   
A way to explicitly generate $Y_i$ is to fix a sequence of i.i.d. random variables $Z_i$ uniformly distributed on the interval $[0,1]$, independent of $X_i$ and of $\xi_i$, and to set
\[Y_i =\I_{Z_i\leq \eta(X_i)}.\]

Now fix $x\in \Omega$ and define auxiliary i.i.d. Bernoulli random variables
\[Y_i^{\prime}=\I_{Z_i\leq \eta(x)},\]
all having the same probability of success $\eta(x)$. They are independent of $X_1,\ldots,X_n$ and of $\xi_1,\ldots,\xi_n$, and are  correlated with the labels $Y_1,Y_2,\ldots,Y_n$.

Let $X_{(1)}(x)$ denote the nearest neighbour to $x$ within the sample $(X_1,\ldots,X_n)$ chosen, if need be, with the help of the auxiliary variables $(\xi_i)$. Denote $Y_{(1)}(x)$ and  $Y^\prime_{(1)}(x)$ the random labels as above assigned to $X_{(1)}(x)$. In particular, $Y_{(1)}(x)$ is the predictor made by the $1$-NN classifier for the point $x$ on the random input $(X_1,\ldots,X_n,Y_1,\ldots,Y_n)$.

The following is Lemma 5.2 in \cite{DGL} in the particular case $k=1$. (It was stated for the Euclidean space, but the nature of the distance in the domain does not matter.)

\begin{lmm}
  $\E \left\vert Y_{(1)}(x) - Y^\prime_{(1)}(x)\right\vert \leq \E \left\vert \eta(x)-\eta(X_{(1)}(x))\right\vert$.
  \label{l:eta}
\end{lmm}

\begin{proof}
  Conditioning on the sample $X_1,\ldots,X_n,\xi_1,\ldots,\xi_n$, we end up with two Bernoulli random variables of the form $Y^\prime=\I_{Z\leq p}$ and $Y=\I_{Z\leq p_1}$ and just need to upper bound $\E\abs{Y-Y^\prime}=P[Y\neq Y^\prime]$ by $\abs{p-p_1}$. 
\end{proof}

\begin{lmm}
  As $n\to\infty$, $\E \left\vert \eta(X)-\eta(X_{(1)}(X))\right\vert\to 0$.
  \label{l:zero}
\end{lmm}

\begin{proof}
  This is the second part of the proof of Theorem 2.2 in \cite{CG} starting with p. 344$_{-3}$, again corresponding to the particular case of fixed $k=1$. The assumption $k\to\infty$ is essential in the first part of the proof, but in the second part, only the assumption $k/n\to 0$ is needed to invoke the Cover--Hart lemma at the very end, subject to a slight adjustment in the proof as indicated to us by Fr\'ed\'eric C\'erou (see Subsection \ref{ss:21}).
\end{proof}

Let now $(X,Y)$ be a random pair following the same law as $(X_1,Y_1)$ and independent of the rest of random variables.
Define the misclassification error of the $1$-NN classifier
\[\ell_n = P[Y_{(1)}(X)\neq Y],\]
and introduce the misclassification error of the auxiliary classifier making for the point $x$ the prediction $Y^\prime_{(1)}(x)$:
\[\ell^\prime_n = P[Y^\prime_{(1)}(X)\neq Y].\]

\begin{lmm}[Cf. top of the p. 70 in \cite{DGL} in the Euclidean case]
  $\E\left\vert \ell_n - \ell^\prime_n\right\vert \to 0$ when $n\to\infty$.
\end{lmm}

\begin{proof} Using Lemmas \ref{l:eta} and \ref{l:zero},
  \begin{align*}
    \E\left\vert \ell_n - \ell^\prime_n\right\vert & \leq P[Y^\prime_{(1)}(x) \neq Y_{(1)}(X)]\\
    &\leq \E  \left\vert \eta(X)-\eta(X_{(1)}(X))\right\vert \\
    &\to 0.
    \end{align*}
\end{proof}

It remains only to estimate the error of the auxiliary classification rule $Y^\prime_{(1)}(X)$, which is straightforward, as now we deal with the probability that two independent identically distributed Bernoulli variables take different values:
\begin{align*}
  \ell^\prime_n & = P[Y^\prime_{(1)}(X)\neq Y]\\
  &=\E\left(2\eta(X)(1-\eta(X))\right) \\
  &\leq 2\E\min\{\eta(X),1-\eta(X)\} \\
  &= 2\ell^{\ast}.
\end{align*}
(Cf. Theorem 5.1 in \cite{DGL}.)

Thus, asymptotically, the classification error of the $1$-NN rule in a complete separable metric space satisfying any one of the equivalent conditions in Theorem \ref{th:equivalences} (for instance, being sigma-finite dimensional in the sense of Nagata if you will) is bounded by twice the Bayes error of the labelled data distribution:
\[\ell^{\ast}(\mu,\eta)\leq\limsup_{n\to\infty}\ell_n\leq 2\ell^{\ast}(\mu,\eta).\]

\end{document}